\documentclass[conference]{IEEEtran}
\IEEEoverridecommandlockouts
\usepackage{amsmath,amssymb,amsfonts}
\usepackage{algorithmic}
\usepackage{algorithm}
\usepackage{graphicx}
\usepackage{textcomp}
\usepackage{siunitx}
\usepackage{xcolor}
\usepackage{booktabs}
\usepackage{array}
\usepackage{hyperref}
\usepackage{ragged2e}
\usepackage{multirow}
\usepackage{makecell}
\usepackage{soul}
\usepackage[skip=10pt,font=normalsize]{caption}
\usepackage{subcaption}
\usepackage[
    backend=biber,
    style=ieee,
    sorting=none
]{biblatex}

\begin{document}

\title{
    Advancing Accessible Underwater Robotics: The Mini-Girona I-AUV at RAMI 2025
    \thanks{
        Research reported in this publication was supported by the University of Girona's Patronat EPS and by the European Union under project number [619789-EPP-1-2020-1-ES-EPPKA1-JMD-MOB].
    }
    \thanks{
        This preprint has not undergone peer review (when applicable) or any post-submission improvements or corrections. The Version of Record of this contribution is published in \textit{ROBOT 2025: Advances in Robotics Proceedings of the Eight Iberian Robotics Conference}, and is available online at \url{https://doi.org/10.1007/978-3-032-30156-7_40}.
    }
}

\author{
    Taqi Hamoda, Bilal Ahmed, Deborah Ele-Ojo, Thi Tran Ha Bao, Adel Saidani, Mazen Elgabalawy, \\ Alaaeddine Chaarani, Sebastian Realpe, Patryk Cie\'{s}lak, Pere Ridao, Narc\'{i}s Palomeras, Nuno Gracias
}

\maketitle

\begin{abstract}

The Mini-Girona Intervention Autonomous Underwater Vehicle (I-AUV) represents an advancement in accessible underwater robotics, designed to bridge the gap between costly, specialized research AUVs and basic Remotely Operated Vehicles (ROVs). Developed with a focus on affordability and usability, the Mini-Girona, priced at approximately \$50,000, integrates advanced components such as a 5-DOF manipulator arm, stereo vision, and AI-driven processing for autonomous navigation and intervention tasks. This paper presents the design and development of the Mini-Girona, detailing its performance during the RAMI 2025 student competition. Despite challenges such as thermal management issues and restricted team access, the Mini-Girona achieved second place overall, excelling in vision-based perception and intervention tasks. This work highlights the platform's potential as a tool for underwater robotics research and education, fostering innovation in real-world underwater applications.

\end{abstract}


\section{Introduction}

The robotics field has significantly advanced across mobile, aerial, legged, and humanoid platforms \cite{robot_survey}. However, high costs and limited accessibility have impeded underwater robotics development \cite{auv_survey}. High-performance underwater robotic platforms, such as intervention-capable Autonomous Underwater Vehicles (AUVs), are prohibitively expensive, costing hundreds of thousands of dollars, with cheaper alternatives providing only basic functionalities \cite{minigirona}. This financial barrier restricts innovation and advanced research, often limiting it to well-funded institutions or those with restrictive rental agreements \cite{minigirona}.

Therefore, accessible yet capable platforms are crucial for broader research and experimentation in real-world underwater environments. This need is amplified by a shift in AUV design priorities. Traditional torpedo-shaped AUVs prioritize speed and long-distance cruising, but their underactuated nature complicates control in complex environments \cite{auv_survey, minigirona}. Recent AUV design trends favor agility, hovering, and subsea residency over speed, enabling precise maneuverability for close-proximity tasks. This shift from broad-survey designs underscores the growing importance of Intervention AUVs (I-AUVs) \cite{auv_survey, girona500}.

\begin{figure}[t!]
    \centering
    \includegraphics[width=\columnwidth]{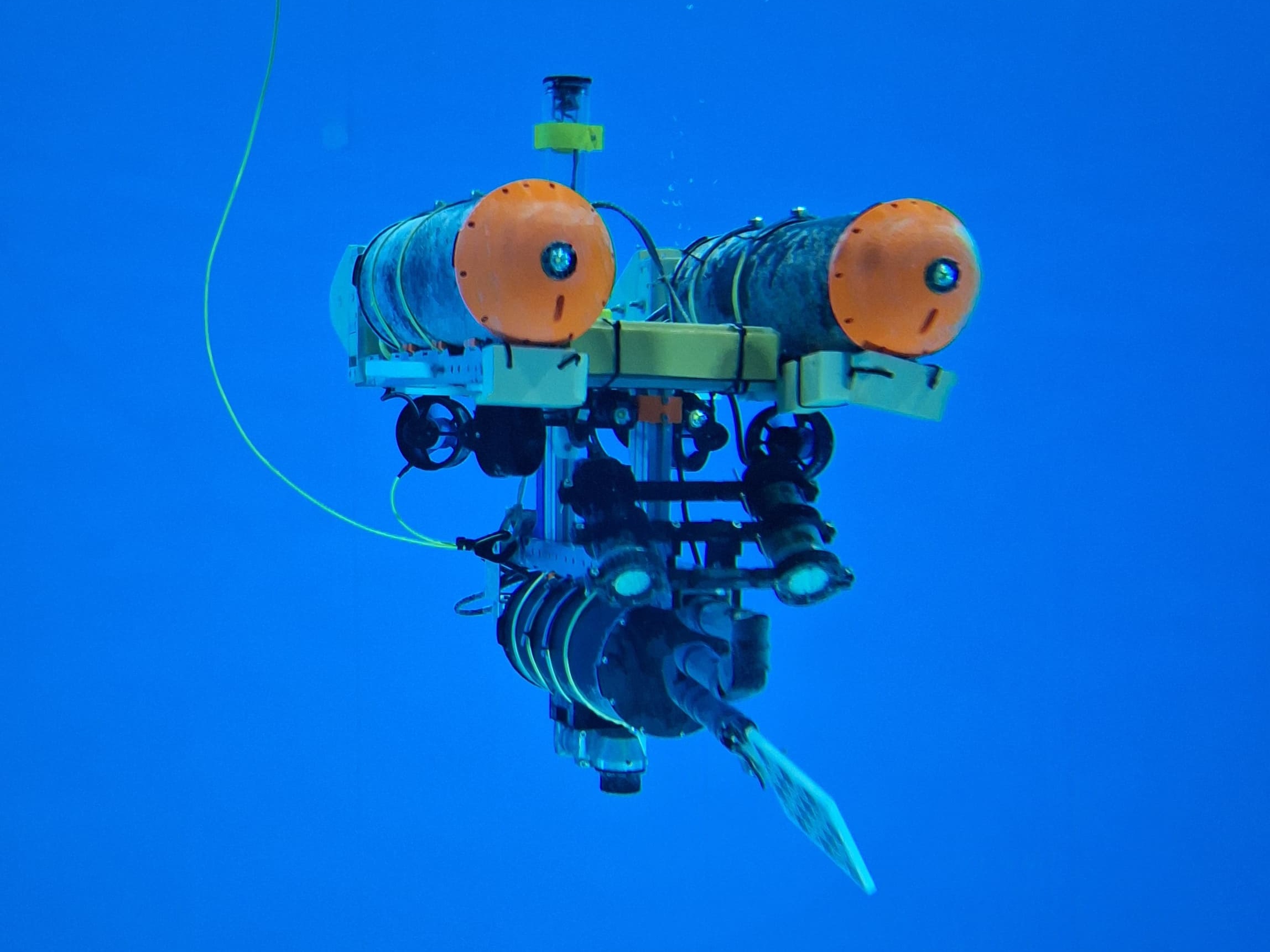}
    \caption{The Mini-Girona I-AUV conducting an intervention task in the CIRS lab's pool.}
    \label{fig:minig}
\end{figure}

To address this need for accessible platforms, the \textbf{Mini-Girona} I-AUV (Fig.~\ref{fig:minig}) was developed. Inspired by the successful Girona 500 I-AUV \cite{minigirona, girona500}, its core philosophy prioritizes affordability and usability without compromising essential functionalities, positioning it as a crucial bridge between basic Remotely Operated Vehicles (ROVs) and prohibitively expensive, specialized research AUVs \cite{minigirona}. This publication provides a comprehensive presentation of the Mini-Girona AUV's design, its development process, and the  algorithms developed and deployed for autonomous operation. It also presents a detailed evaluation of its performance and the results achieved during its participation in the Robotics for Asset Maintenance and Inspection (RAMI) 2025 competition \cite{metricsprojectRAMICall} organized as part of the EU Metrics project. The subsequent sections will guide the reader through the detailed discussion of these aspects.

\section{The Mini-Girona}

\begin{table*}[t!]
    \centering
    \caption{Mini-Girona AUV Specifications}
    \label{tab:mini-girona_specs_rearranged}
    \begin{tabular}{@{} l l l l @{}}
        \toprule
        \multicolumn{2}{c}{\textbf{General}} & \multicolumn{2}{c}{\textbf{Actuators \& Sensors}} \\
        \cmidrule(r){1-2}\cmidrule(l){3-4}
        Weight & 30-35 kg & Thrusters & 6 Blue Robotics T200 thrusters \\
        Dimensions & 80 x 75 x 75 cm & Control DOF & Surge, Sway, Heave, Yaw \\
        Payload Capacity & 10 kg & Manipulator Arm & Reach Alpha 5 DOF by Blueprint Lab \\
        Depth Rating & 50 m & Manipulator Reach & 400 mm \\
        Autonomous Operation & Up to 11 hours & DVL & Nortek Nucleus 1000 DVL \\
        Main Processor & Intel NUC PC (Linux, ROS) & GPS & Quectel L86-M33 GPS Module \\
        AI Processing Unit & Jetson Nano Orin 8 GB & Imaging Sonar & Tritech Miniking Sonar \\
        Microcontroller & STM32 NUCLEO-32 & Sonar Range & 100 meters \\
        Battery Management System & EMUS BMS & Stereo Vision System & 2 Basler ace 2 cameras \\
        Underwater Comms & Tether & Camera Resolution & 1920 x 1200 \\
        Emergency Switch & Magnet-based contactless switch & Camera Frame Rate & 5 FPS \\
        \bottomrule
    \end{tabular}
\end{table*}

The Mini-Girona AUV is an autonomous underwater vehicle developed for research, inspection, and intervention, building upon the Girona 500 AUV's design principles. It offers enhanced accessibility to underwater robotics due to its significantly lower cost of approximately \$50,000, in contrast to the Girona 500's \$400,000, aiming to broaden accessibility and participation in underwater robotics for research and education \cite{minigirona}. 

The Mini-Girona platform features a Blueprint Lab Reach Alpha 5-DOF manipulator arm with a \SI{400}{\milli\meter} reach, enabling tasks such as sample collection and object manipulation. For perception, the Mini-Girona is equipped with two Basler ace 2 cameras in a stereo configuration, supporting object identification and 3D reconstruction. Navigation is facilitated by a Nortek Nucleus 1000 Doppler Velocity Log (DVL), a Tritech Miniking mechanical scanning imaging sonar, and an NMEA GPS module. The vehicle integrates a Jetson Nano Orin 8GB board for real-time deep learning and AI processing, alongside an Intel NUC PC for control and sensor communication. A summary of the specifications is presented in table \ref{tab:mini-girona_specs_rearranged}.

\begin{figure}[t!]
    \centering
    \includegraphics[width=\columnwidth]{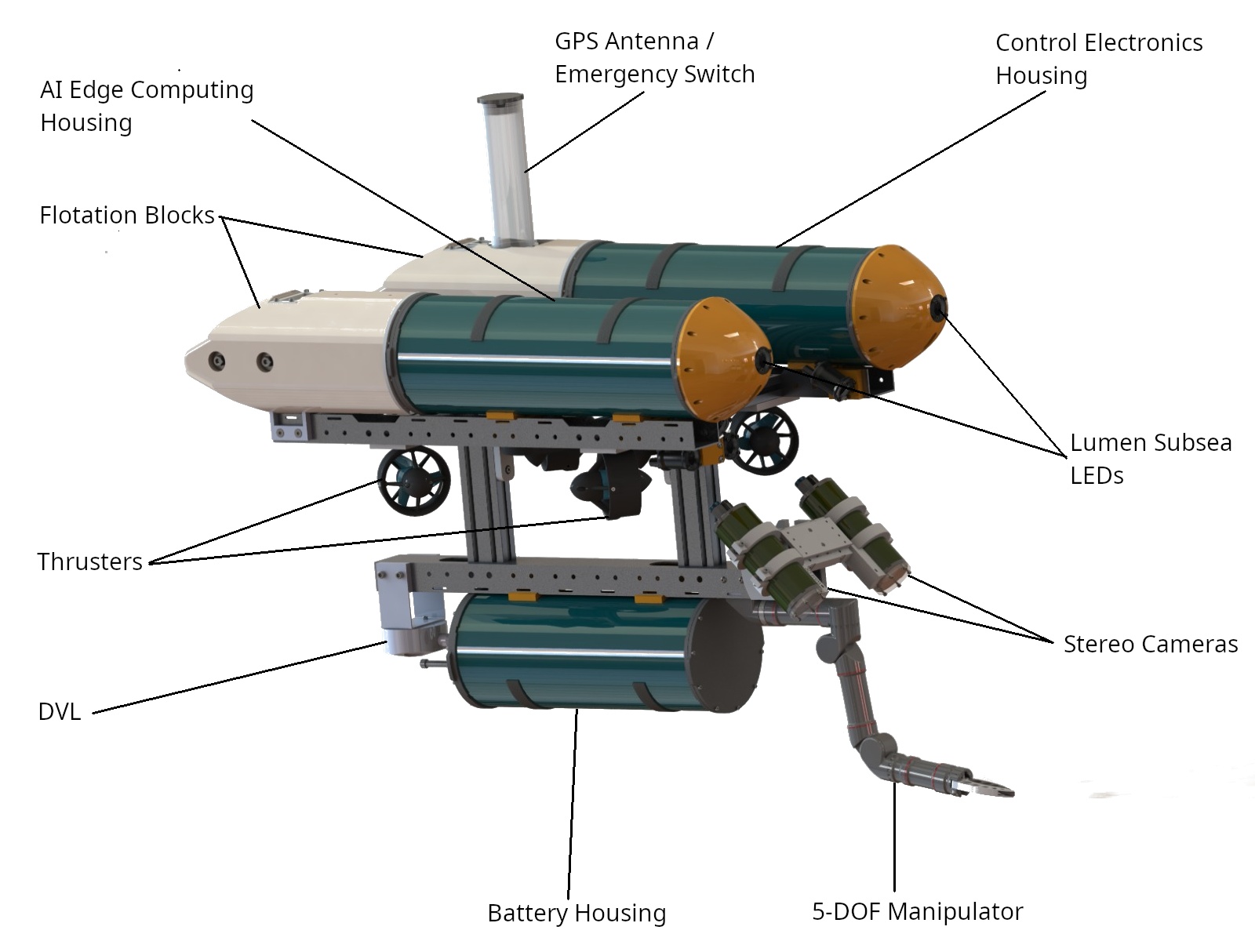}
    \caption{An annotated diagram of the Mini-Girona's components.}
    \label{fig:minig_annotated}
\end{figure}

\subsection{COLA2 Control Architecture}

The Mini-Girona utilizes the COLA2 (Component Oriented Layer-based Architecture for Autonomy) hierarchical control architecture for autonomous operation \cite{minigirona, cola2}. COLA2 comprises three layers:
\begin{itemize}
    \item \textbf{Reactive Layer:} Manages real-time sensor/actuator interactions and executes fundamental robot behaviors, integrating Reinforcement Learning for enhanced adaptability.
    \item \textbf{Execution Layer:} Translates high-level mission plans into low-level commands for the Reactive Layer, employing an Architecture Abstraction Component and a Petri Net Player for mission execution.
    \item \textbf{Mission Layer:} Defines high-level goals and tasks, supporting both predefined mission plans via Mission Control Language and dynamic mission management for real-time modification and replanning.
\end{itemize}

\section{The RAMI Competition}

Underwater robotics competitions significantly advance technology and develop expertise by providing practical experience with real-world complexities like variable salinity and turbidity, often not covered in academic curricula \cite{competitions_survey}. Europe's first such competition, SAUC-E, launched in 2006, drawing inspiration from the 1997 US RoboSub \cite{sauce_to_erl}. Since 2010, the Center for Maritime Research and Experimentation (CMRE) has hosted SAUC-E, focusing on Autonomous Underwater Vehicles (AUVs) for harbor missions. SAUC-E's success led to CMRE's creation of the ERL Emergency events (2017 -- 2019), which continued to emphasize multi-domain challenges and autonomous decision-making \cite{sauce_to_erl}. Since 2021, ERL has transitioned to RAMI, under the METRICS EU Horizon 2020 project, concentrating on quantitative evaluation of robot performance for Inspection \& Maintenance (I\&M) tasks in high-risk environments \cite{erl_to_rami, rami}.

\subsection{RAMI Competition Structure}

The RAMI Marine Robots competition is an international challenge designed to advance autonomous underwater robotics for critical infrastructure inspection and maintenance tasks. RAMI 2025, held in the CMRE marina in La Spezia, Italy, focused on deploying autonomous underwater vehicles (AUVs) in realistic, high-risk scenarios where human intervention is impractical. The competition evaluated teams on autonomous navigation, data acquisition, object detection, and manipulation, emphasizing safety and environmental sustainability. The competition area measured $50 \times 25$ \si{m}, with a depth ranging from $4$ to $6$ \si{m} and underwater visibility between $1$ and $3$ \si{m} \cite{rami}.

The competition simulated an emergency situation where critical undersea infrastructure (CUI) for data communication has been badly damaged. Robotic teams deployed AUVs to assess damage, locate the compromised pipe, and close a valve to prevent further environmental and structural harm. Competing AUVs executed the following Task Benchmark Missions (TBMs):

\begin{itemize}
    \item \textbf{TBM-1: Pipeline Area Inspection} \\
    AUVs navigated to the incident site using waypoint data from UAVs via CATL messages. Tasks included mapping colored buoys, passing through a designated gate, and inspecting a designated structure.
    \item \textbf{TBM-2: Intervention on Pipeline Structure} \\
    AUVs followed the main pipeline, detected damage markers (red or black), reported their locations, and performed intervention tasks, such as turning a valve on the manipulation console by \ang{90} and maintaining direct contact with the damaged area for one minute. The mission concluded with the retrieval and surfacing of a ring-pole.
    \item \textbf{TBM-3: Comprehensive I\&M Mission} \\
    This comprehensive task combined TBM-1 and TBM-2, requiring autonomous or semi-autonomous inspection, localization, and intervention to resolve the emergency.
\end{itemize}

\subsection{Previous Notable Achievements}

The VICOROB-UdG Team has participated in several European student AUV competitions. Their first AUV, \textbf{ICTINEU}, won SAUC-E 2006 seven months after its creation \cite{ictineu_team}. Four years later, the team introduced the \textbf{SPARUS} AUV which won SAUC-E 2010 within six months of being operational with subsequent refinements leading to the \textbf{SPARUS II} AUV, which Iqua Robotics later commercialized \cite{sparus_team, sparus2}.

The UNIFI Robotics Team from the University of Florence has been a consistent presence in underwater robotics competitions since 2011, participating in ERL Emergency in 2017, 2018 and 2019, and RAMI in 2022, 2023, and 2025 with their AUV, the \textbf{FeelHippo} \cite{unifi_team2020, unifi_team2023}. In RAMI 2023, the UNIFI Robotics Team excelled, securing first place in multiple categories, including pipeline inspection and intervention, and complete plant missions, earning the "Best team SAUC-E award" \cite{unifi_team2023}.

Team ERGO from the University of Pisa participated in RAMI 2023 with their \textbf{Zeno} vehicle \cite{pisa_team}. Their object detection and classification algorithm, which utilizes color enhancement, deep learning (Faster R-CNN), and color/shape-based methods, achieved second place in the 1st Marine Cascade Campaign, the virtual RAMI competition, with a mean Average Precision (mAP) of approximately 92\% on provided datasets \cite{pisa_team}.

\section{Methodology}

A physics-based simulator for underwater robotics research, Stonefish \cite{stonefish}, was used \cite{stonefish} to validate the developed algorithms. A detailed model of the Mini-Girona robot, including all onboard sensors, was implemented in Stonefish, replicating the RAMI competition scenario to test the developed algorithms before physical deployment. Subsequently, real-world validation was conducted in the CIRS pool with RAMI scenario components (e.g. buoys, pipe, and a custom-built console) placed there to mimic the environment present at the competition (Fig.~\ref{fig:pool_sim_setup}). The robot's hardware, including DVL, stereo cameras, and safety equipment, was extensively tested in the pool, followed by an evaluation of the implemented algorithms on the aforementioned scenario components.

\begin{table}[t!]
    \centering
    \begin{tabular}{c}
        \includegraphics[width=0.3\columnwidth]{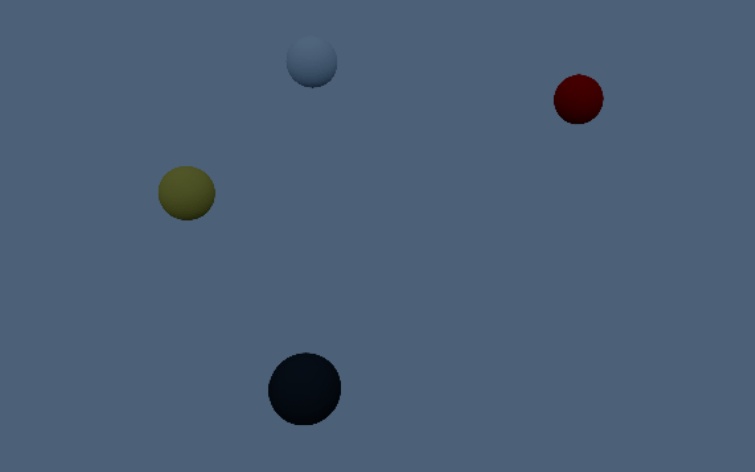}
        \includegraphics[width=0.3\columnwidth]{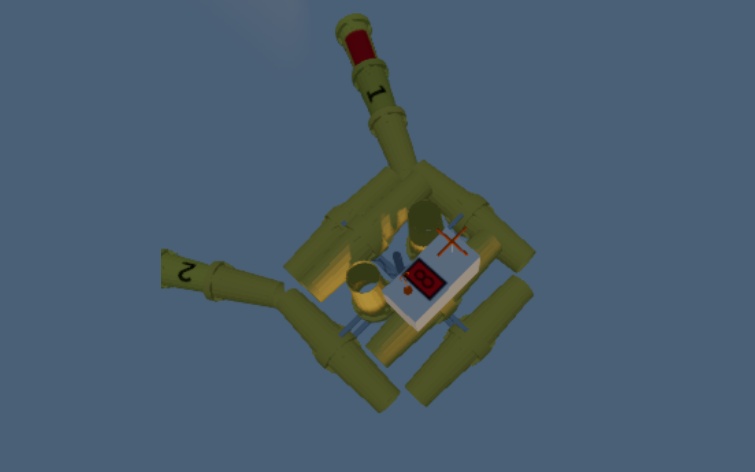}
        \includegraphics[width=0.3\columnwidth]{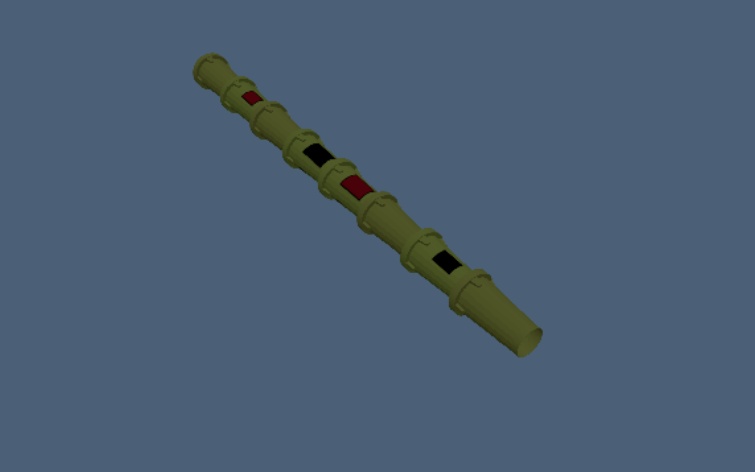} \\
        \includegraphics[width=0.3\columnwidth]{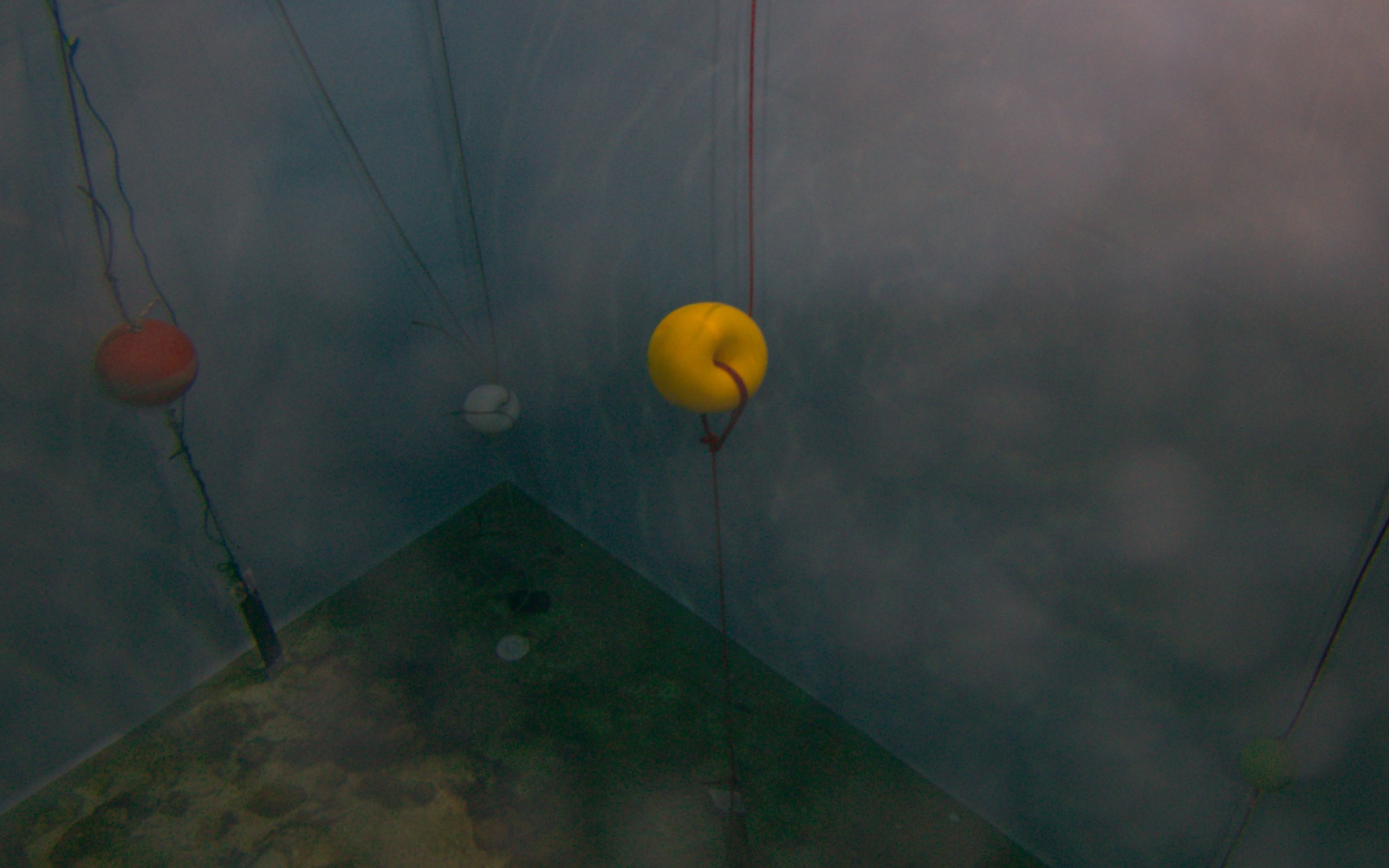}
        \includegraphics[width=0.3\columnwidth]{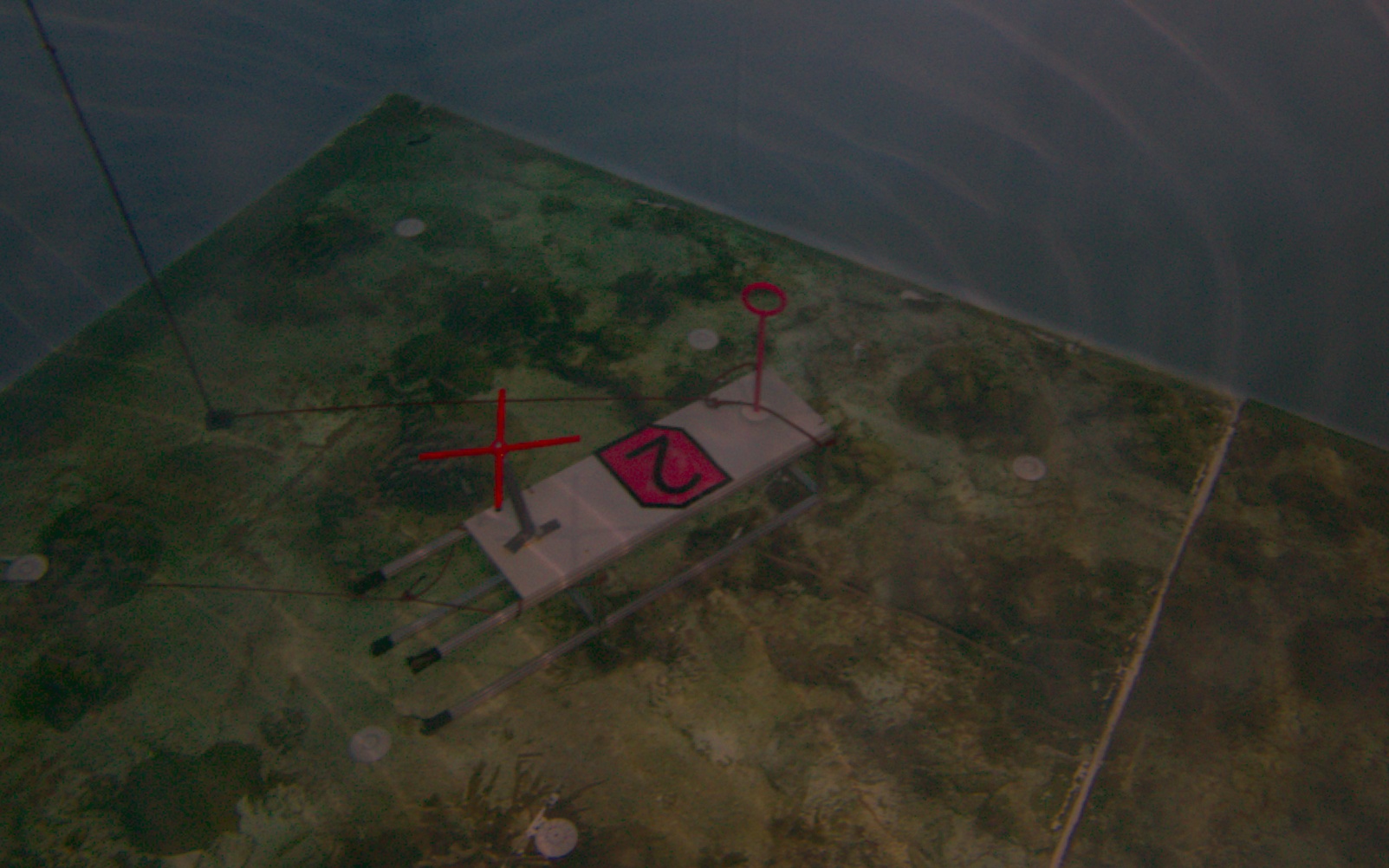}
        \includegraphics[width=0.3\columnwidth]{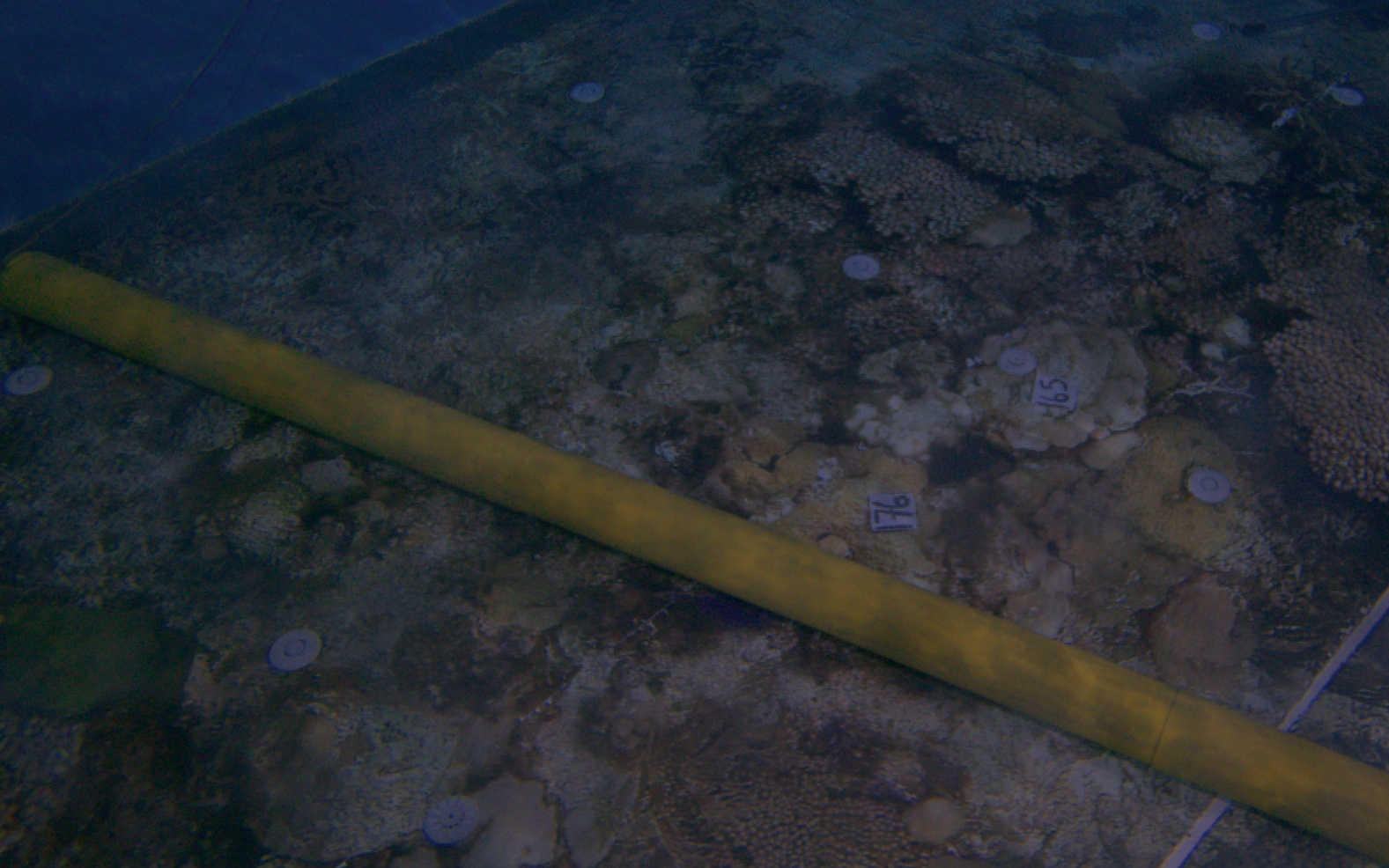}
    \end{tabular}
    \captionof{figure}{Testing setup from the Stonefish simulator and the CIRS pool.}
    \label{fig:pool_sim_setup}
\end{table}

\subsection{Perception}

The Mini-Girona utilizes a stereo camera system to enable 3D reconstruction of underwater scenes. Stereo block matching computes disparities between the rectified stereo images to generate an initial point cloud, refined by a bilateral filter to reduce noise while preserving structural details \cite{bilateral_filter}. DBSCAN clustering is then used to isolate the largest cluster and to eliminate sparse noise \cite{dbscan}. A YOLO-E v11 model is then used for real-time object detection and segmentation, identifying structures such as valves, ring poles, damage markers, pipes, and buoys \cite{YOLO-E}. These segmentation masks, combined with corresponding point clouds, support precise localization of interaction points.

For valve handle localization, Principal Component Analysis (PCA) determines the principal axis of the point cloud to identify the handle and valve center \cite{PCA}. Pipe following employs PCA to calculate the nearest and farthest points along the principal axis, providing navigational guidance. Finally, for precise for number and color recognition, the Large Language and Vision Assistant (LLaVA) model is used to process cropped regions from YOLO-E v11, enhancing scene understanding and intelligent detection \cite{llava}.

\begin{table*}[t!]
    \centering
    \begin{tabular}{c}
        \includegraphics[width=\columnwidth]{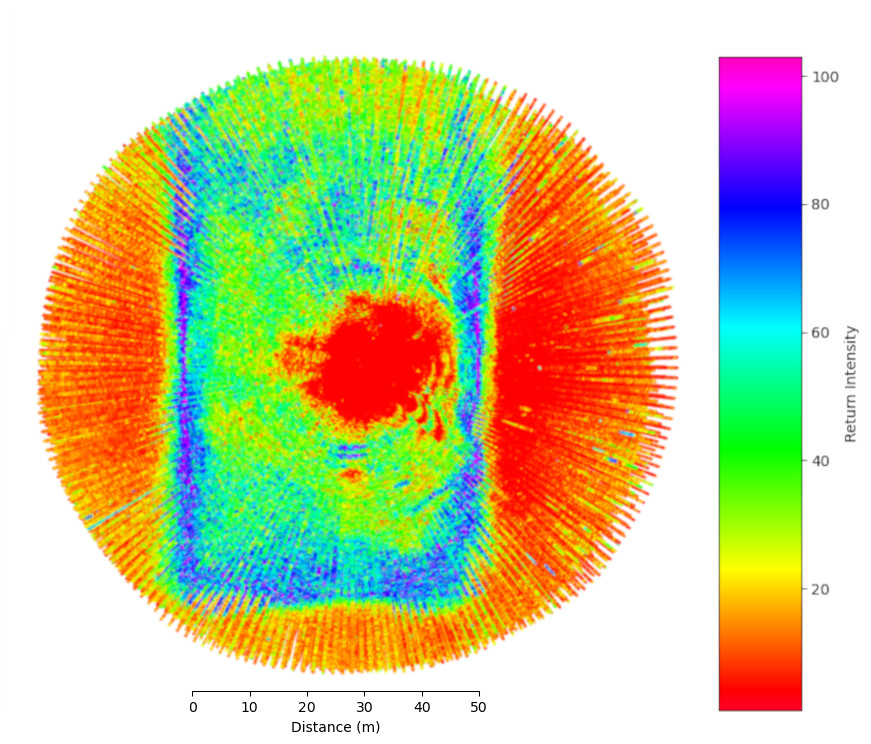}
        \includegraphics[width=\columnwidth]{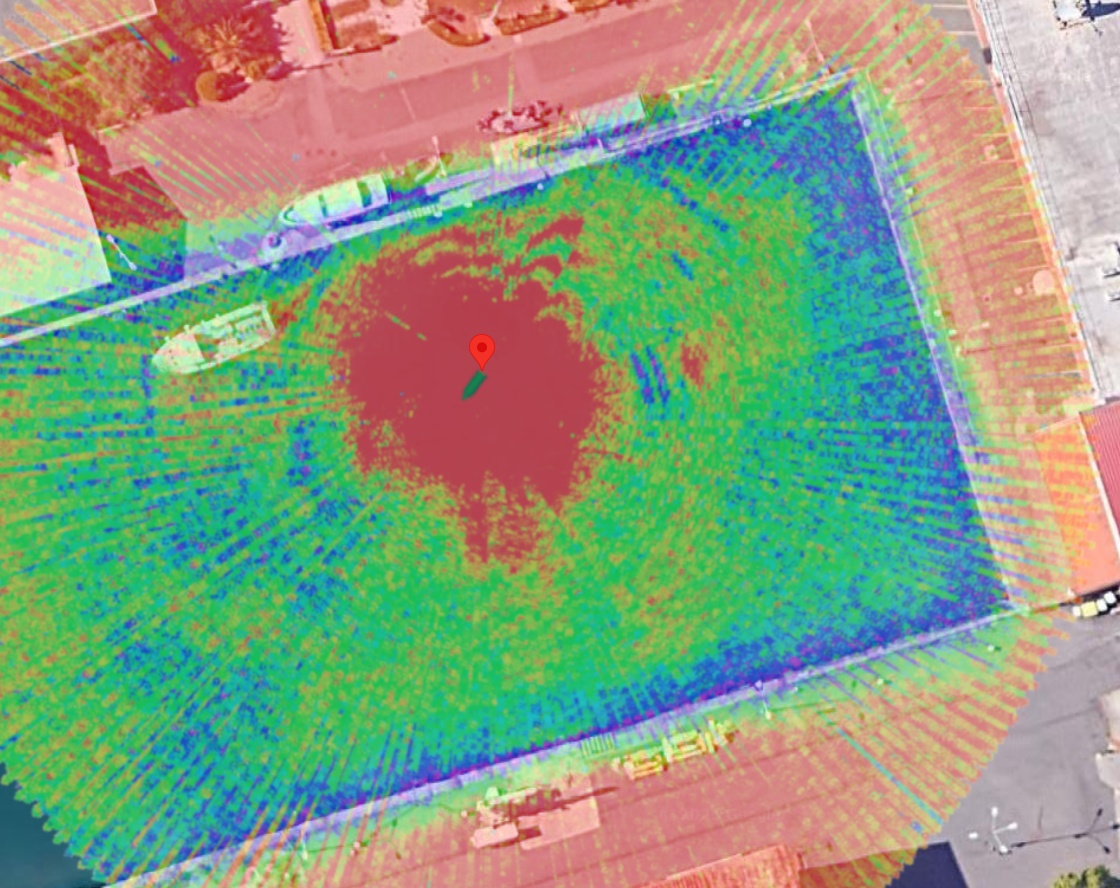} \\
    \end{tabular}
    \captionof{figure}{Acoustic map generated using sonar data collected from the CMRE marina during TBM 3. GPS-alignment was used to accurately overlay the map obtained on satellite imagery of the marina.}
    \label{fig:sonar}
\end{table*}

\subsection{Localization}

The localization framework for the Mini-Girona AUV in structured environments employs a constant velocity model to update the vehicle's state, which includes position, orientation, and velocities. This model integrates data from the DVL, pressure sensor, compass, and sonar, incorporating process noise to ensure robustness. The state update aligns environmental features with the vehicle's orientation for accurate tracking.

An Extended Kalman Filter (EKF) enhances state estimation by fusing measurements from the compass (yaw), depth sensor (z-coordinate), DVL (linear velocities), and inertial measurement unit (angular velocity). Sonar data, expressed as range and bearing, are converted to Cartesian coordinates, and line features are transformed into the vehicle's frame. The EKF associates sonar measurements with environmental features using a statistical distance metric, with the measurement update modeled as:

\begin{equation*}
h(\hat{\mathbf{p}}_c(k), \hat{\mathbf{l}}_n^V(k)) = \rho_n^V - x_c \cos \theta_n^V - y_c \sin \theta_n^V = 0
\end{equation*}

\subsection{Autonomous Navigation and Planning}

AUV navigation and inspection rely on effective path planning and obstacle avoidance. The Rapidly-exploring Random Tree Connect (RRT Connect) algorithm generates paths using a 2D projected map derived from 3D voxel data aggregated from the onboard Miniking sonar. Upon receiving a goal point, the algorithm computes a waypoint list to guide the AUV, avoiding obstacles in the initial map. A proportional controller adjusts linear and angular velocities based on the localization data and the target position, feeding setpoints to the COLA2 architecture for thruster control.

During navigation, newly detected obstacles trigger path replanning from the AUV's current position to avoid collisions. For pipeline inspection, where sonar struggles to detect pipes due to their material density, camera-based point cloud data is fused with sonar data to create a 3D map. This enables safe navigation and 3D reconstruction of the pipeline structure at a constant, slow speed, ensuring the AUV remains oriented toward the structure without collisions.

\subsection{Manipulation and Intervention}

The system supports goal-based and reactive behaviors through controller switching and task-priority control. The joint trajectory controller executes predefined joint configurations, while the joint velocity controller handles task-priority execution, enabling seamless transitions between planned motions (e.g., reaching a pose) and dynamic interventions (e.g., obstacle avoidance).

The end-effector configuration is represented by $\boldsymbol{\sigma}_c(\mathbf{q}) \in \mathbb{R}^{6 \times 1}$ task vector, capturing its pose as a function of joint configuration $\mathbf{q}$. The task error $\tilde{\boldsymbol{\sigma}}_c$ quantifies position and orientation discrepancies, with orientation expressed via quaternion components:
\[
\tilde{\boldsymbol{\sigma}}_c =
\begin{bmatrix}
\eta_{1,d} - \eta_1 \\
w \boldsymbol{\epsilon}_d - w_d \boldsymbol{\epsilon} - \boldsymbol{\epsilon} \times \boldsymbol{\epsilon}_d
\end{bmatrix}
\]
The Jacobian $\mathbf{J}_c(\mathbf{q}) \in \mathbb{R}^{6 \times n}$ relates joint velocities to Cartesian velocities, enabling the task-priority controller to guide the end-effector accurately.

\section{Results}

The Mini-Girona I-AUV participated in the RAMI 2025 competition just three weeks after being operational, in a bid to demonstrate its capabilities across various inspection and intervention tasks in a challenging underwater environment. Despite facing several significant hurdles, the team achieved commendable results, securing second place overall. The team's strong performance in the vision-based perception and intervention tasks led to the Mini-Girona team receiving the following prestigious awards:

\begin{itemize}
    \item TBM 3 - Grand Challenge: \textbf{2nd place}
    \item TBM 2 - Plant Intervention: \textbf{2nd place}
    \item TBM 1 - Plant Inspection: \textbf{3rd place}
    \item Best Team Presentation
    \item Best Student Poster
\end{itemize}

\subsection{AUV Performance}

The sonar system in TBM 3 generated a $75 \times 50$ \si{m} acoustic map of the competition area (Fig.~\ref{fig:sonar}), allowing for robust spatial awareness, navigation, and self-localization. For TBM 1 (Inspection) and TBM 2 (Intervention) tasks, the system produced rectified images for stereo processing, enabling reliable 3D reconstruction via disparity maps. The stereo setup provided crucial 3D depth information for navigation, obstacle avoidance, and object interaction, including pipe following. Additionally, the integrated YOLO-E v11 model identified and localized buoys, pipe markers, the valve, and the ring pole successfully  (Fig.~\ref{fig:stereo_results}). 

\begin{figure}[t!]
    \centering
    \includegraphics[width=\columnwidth]{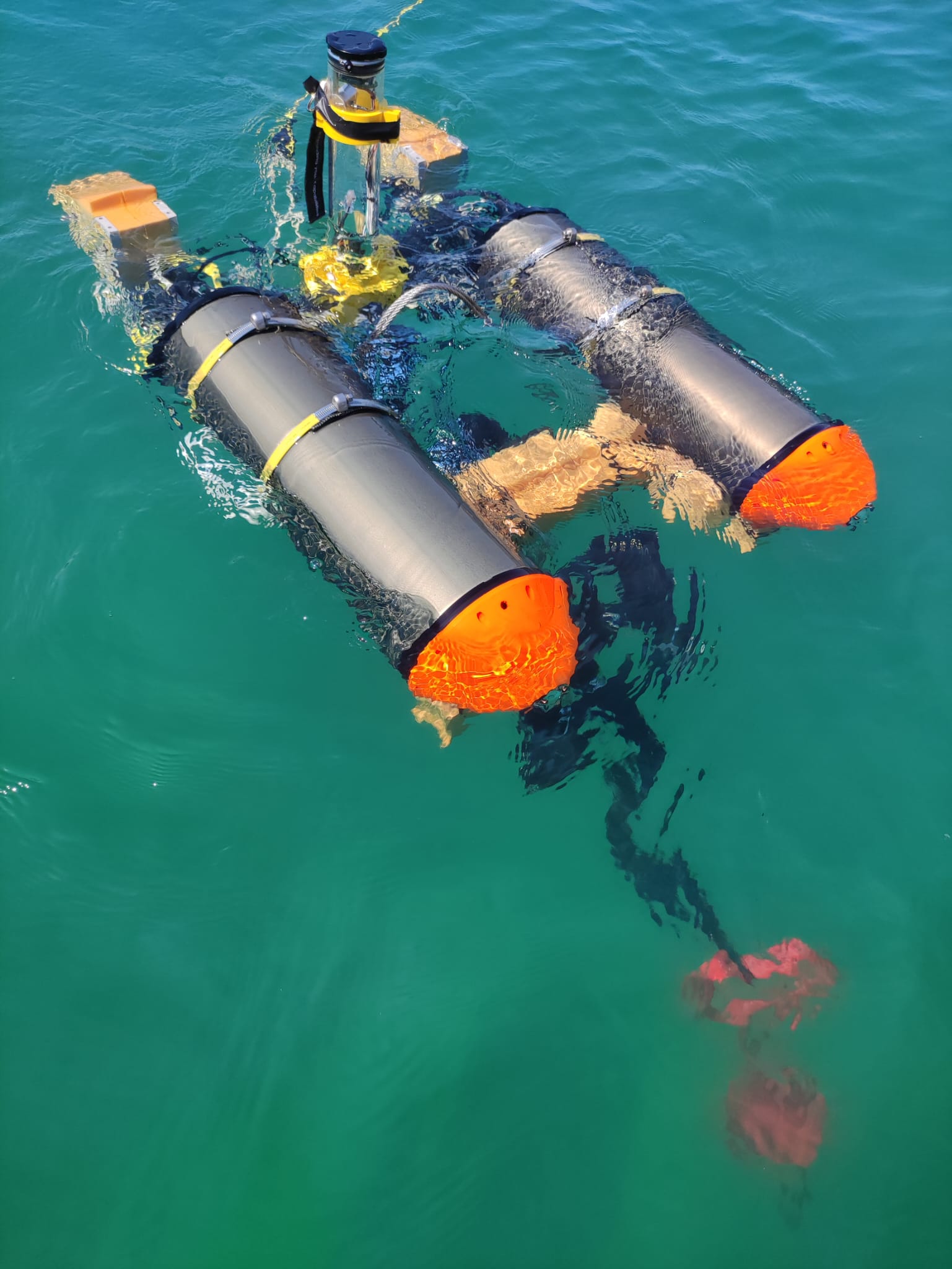}
    \caption{The Mini-Girona surfacing with the pole from TBM-2 grasped by the end-effector.}
    \label{fig:minig_pole}
\end{figure}

The team successfully completed the interventional aspect of TBM 2, notably being the only competitor to turn the valve and maintain physical contact with the pipe marker. This intervention, though completed via teleoperation due to unforeseen challenges, demonstrated the Mini-Girona's manipulator arm's precision and dexterity. As depicted in Figure~\ref{fig:minig_pole}, the Mini-Girona successfully surfaced with the pole grasped, signifying the completion of TBM 2.

\begin{table}[t!]
    \centering
    \begin{tabular}{c}
        \includegraphics[width=0.3\columnwidth]{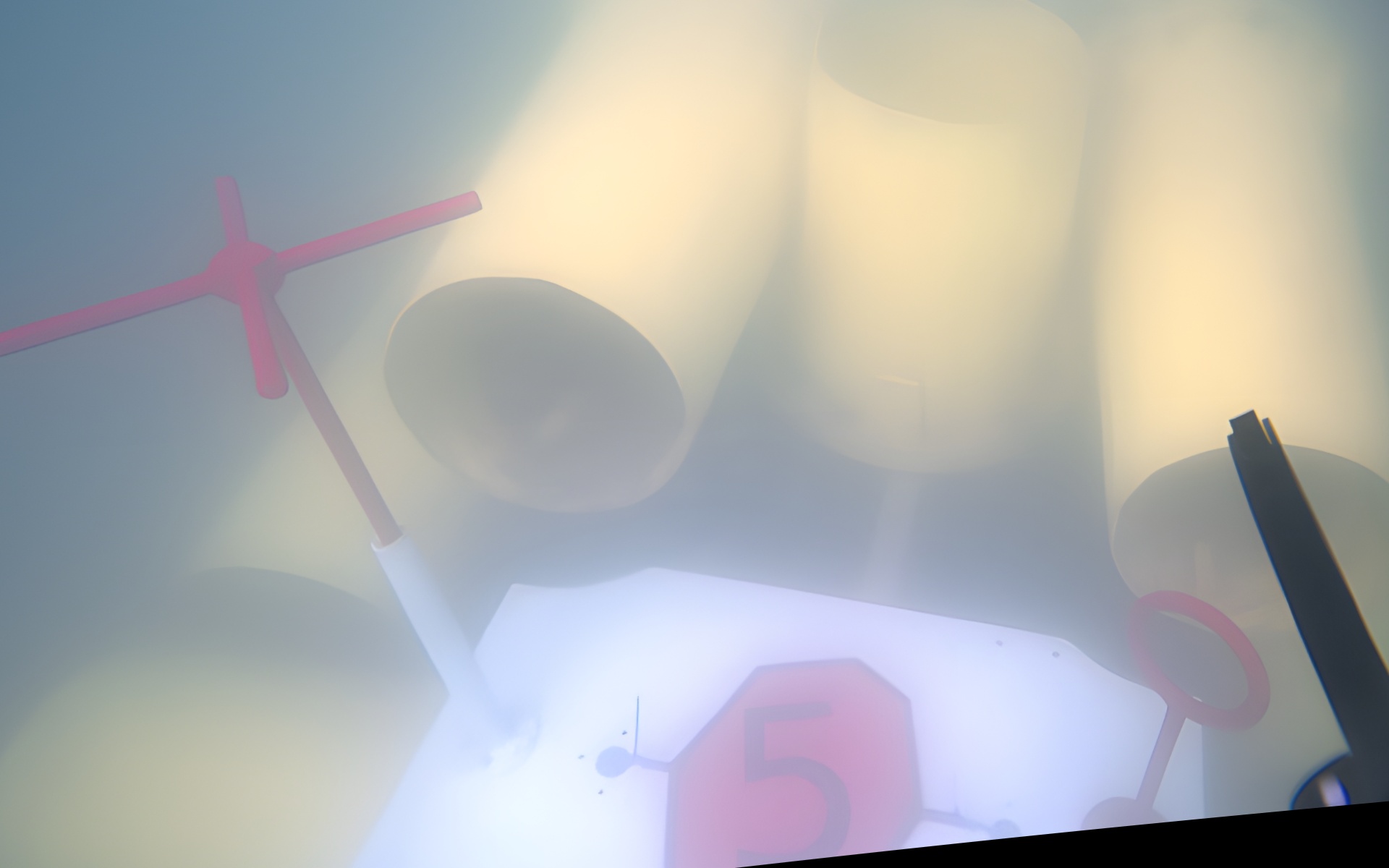}
        \includegraphics[width=0.3\columnwidth]{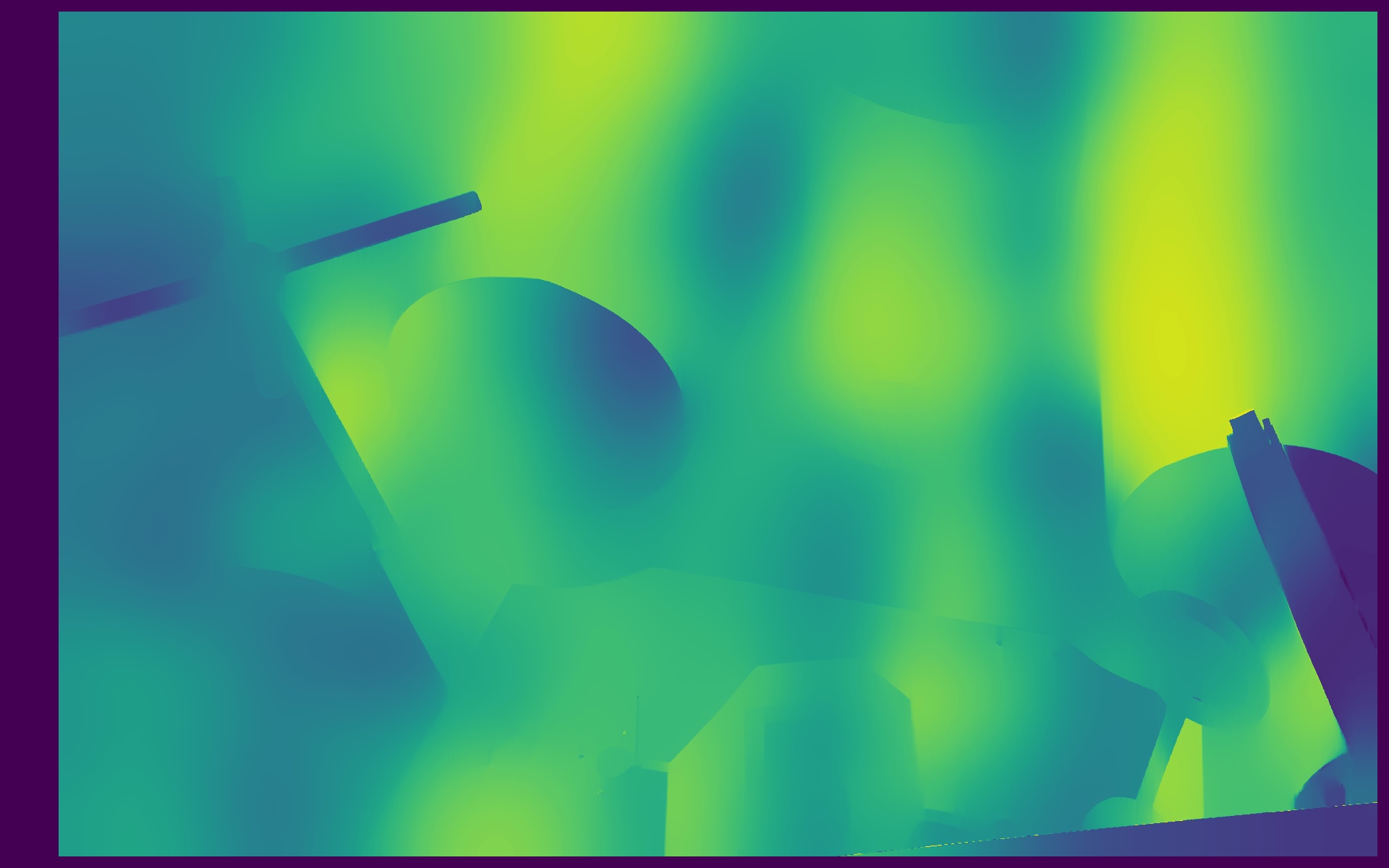}
        \includegraphics[width=0.3\columnwidth]{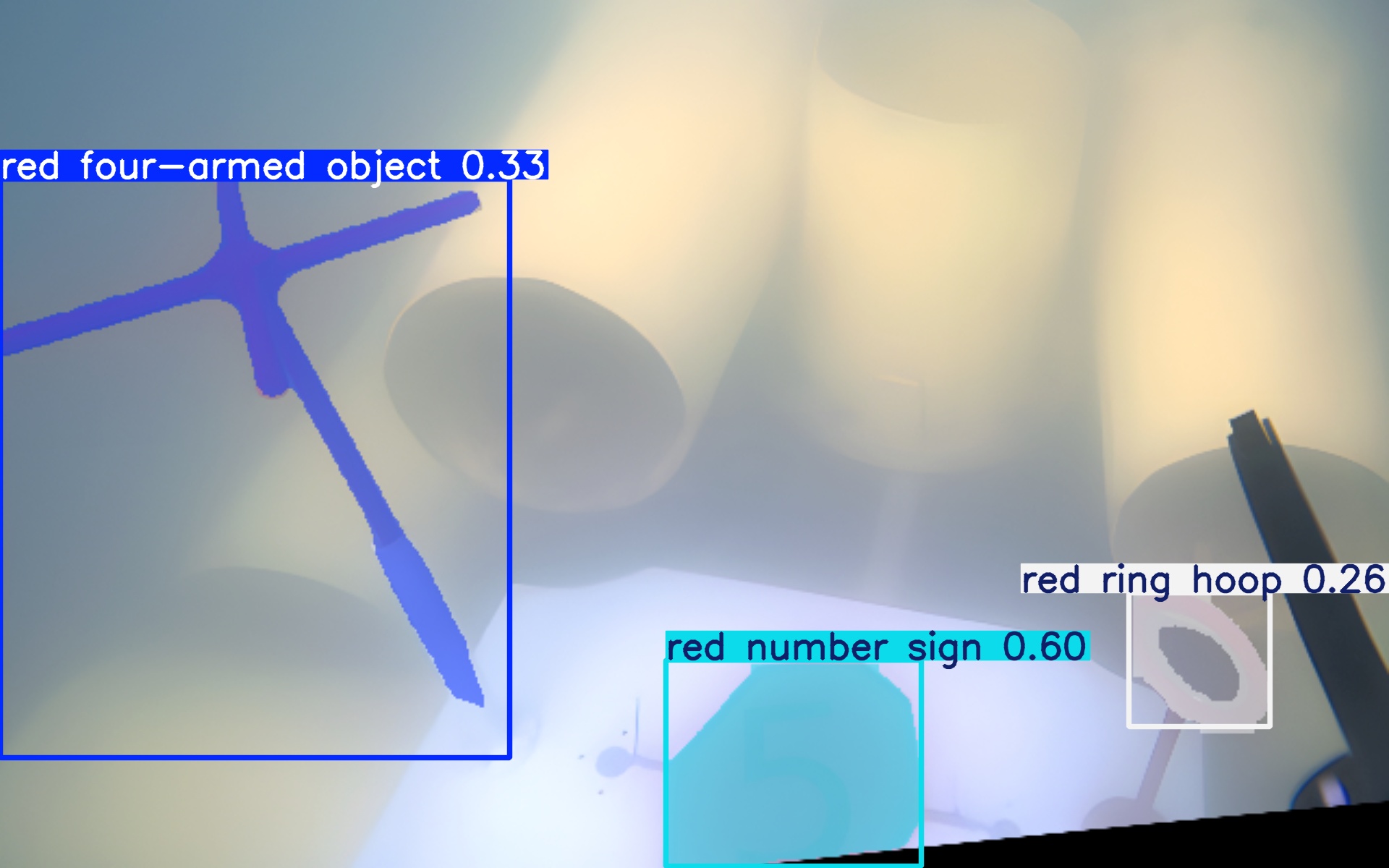} \\
        \includegraphics[width=0.3\columnwidth]{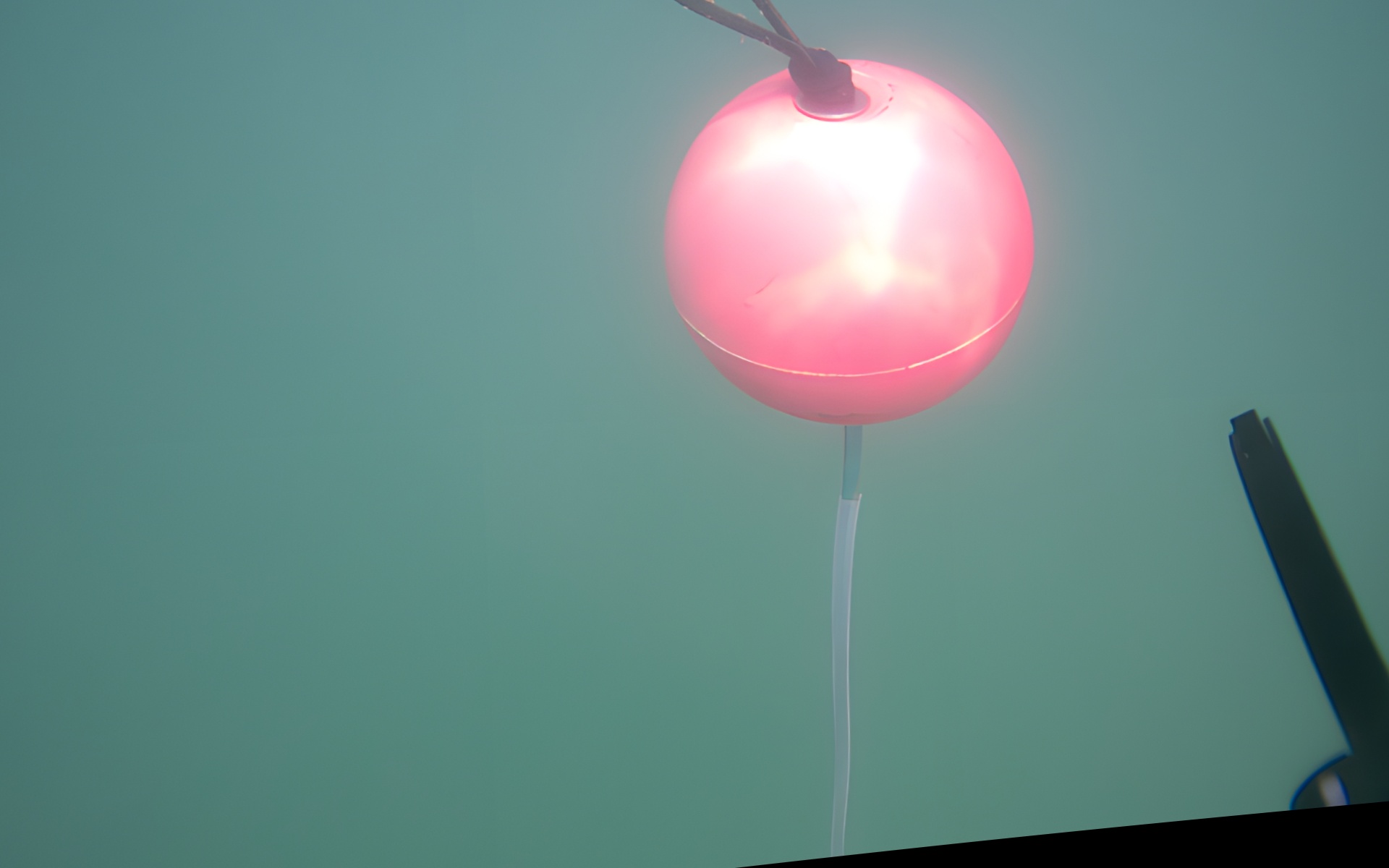}
        \includegraphics[width=0.3\columnwidth]{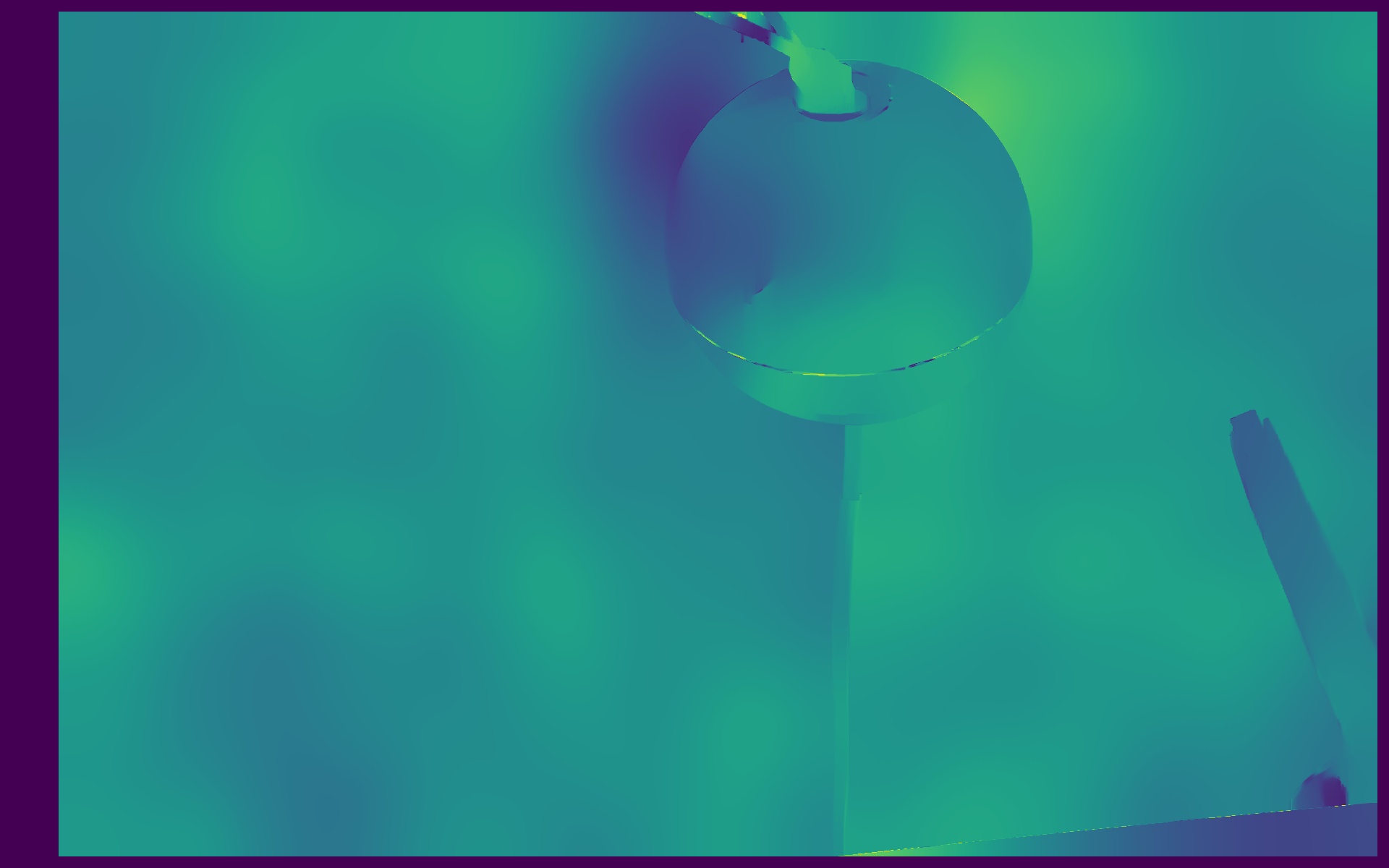}
        \includegraphics[width=0.3\columnwidth]{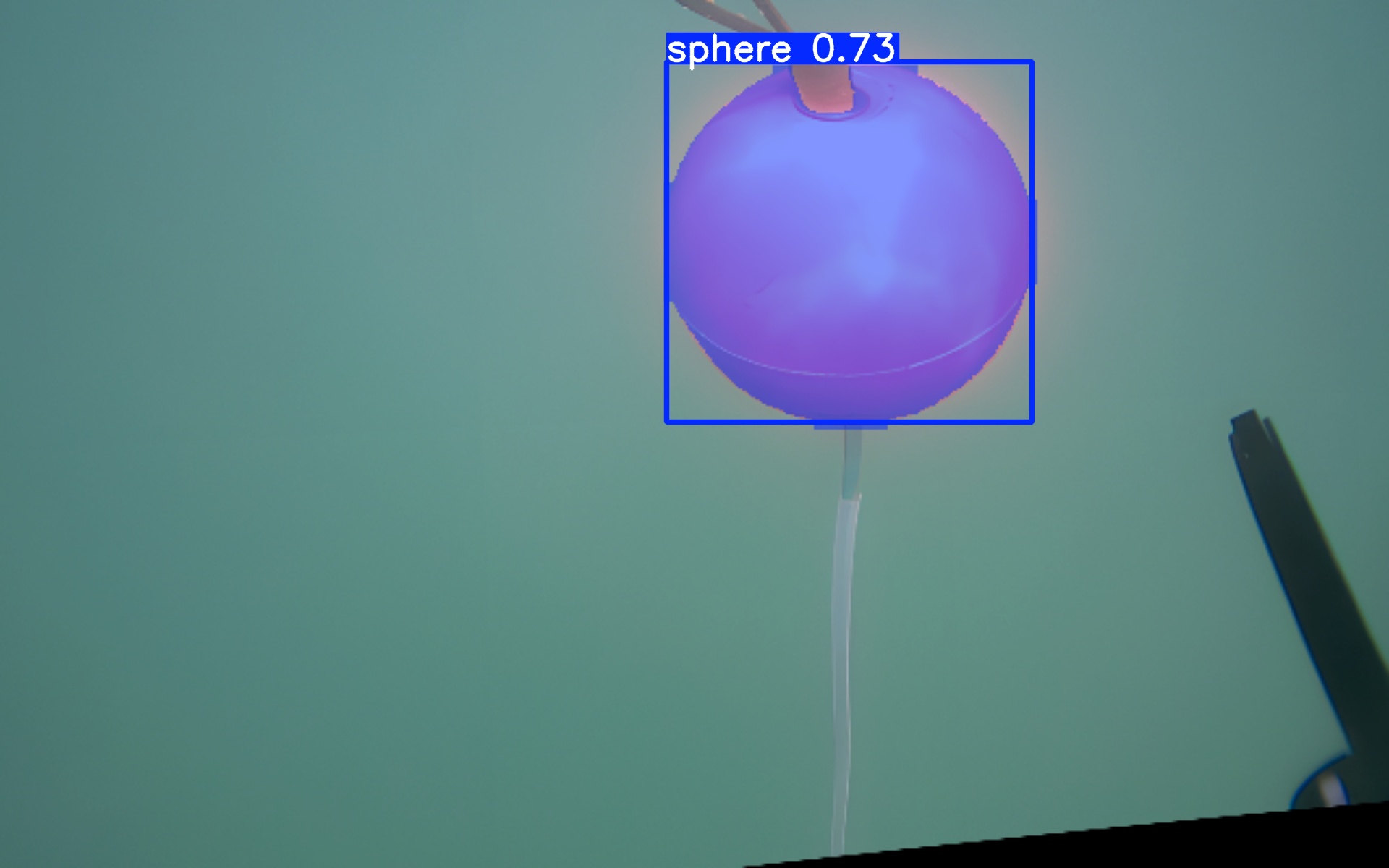} \\
        \includegraphics[width=0.3\columnwidth]{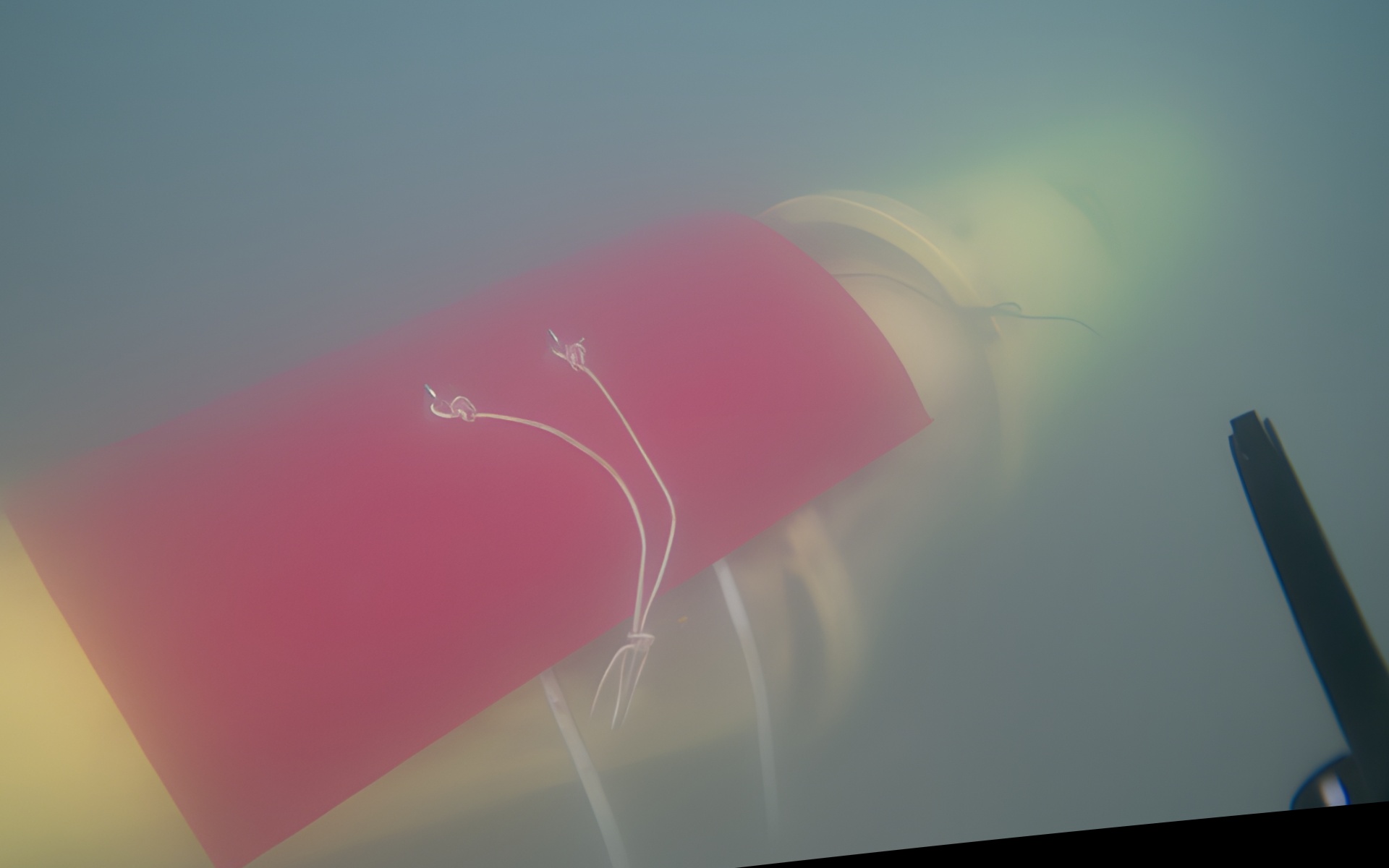}
        \includegraphics[width=0.3\columnwidth]{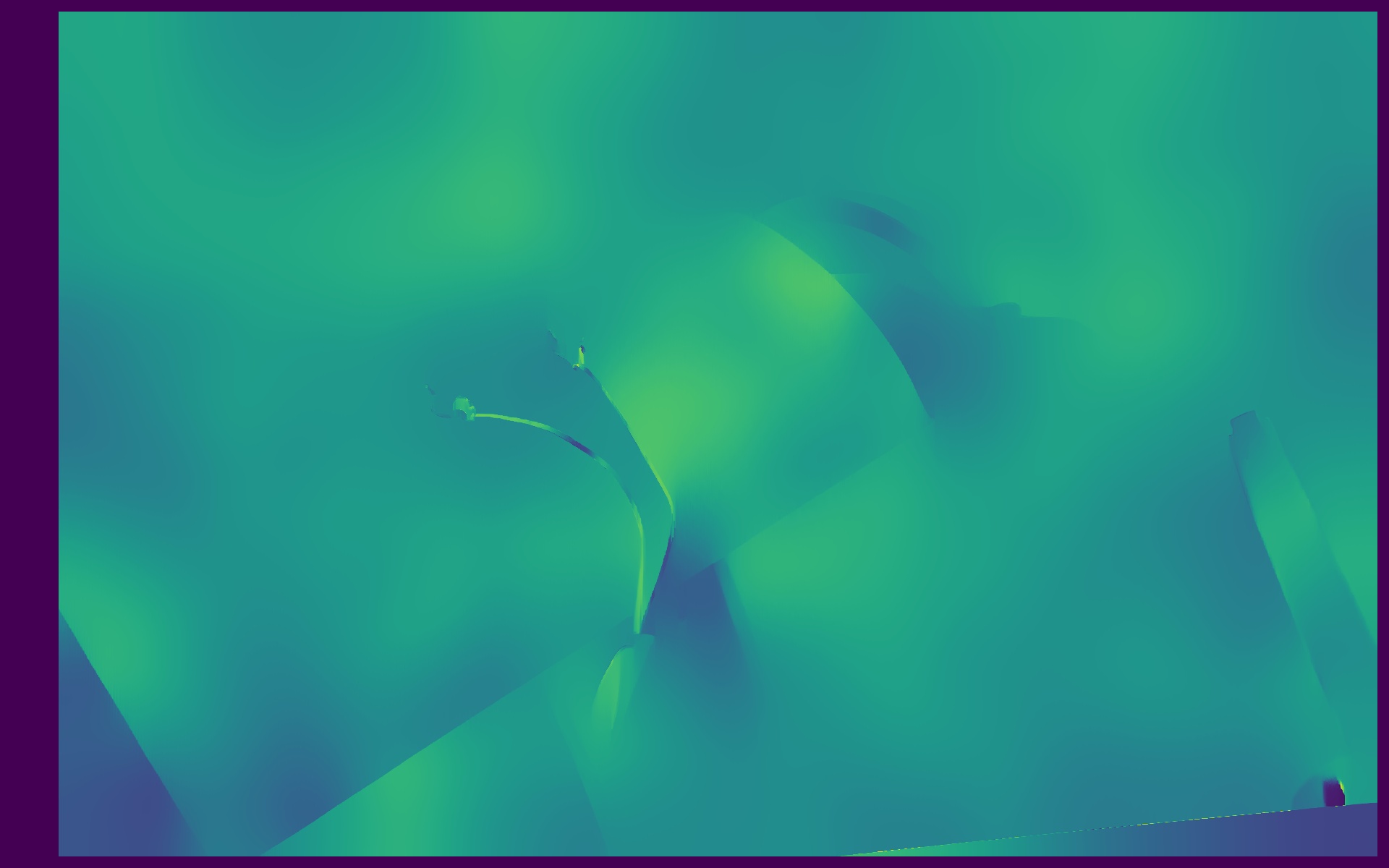}
        \includegraphics[width=0.3\columnwidth]{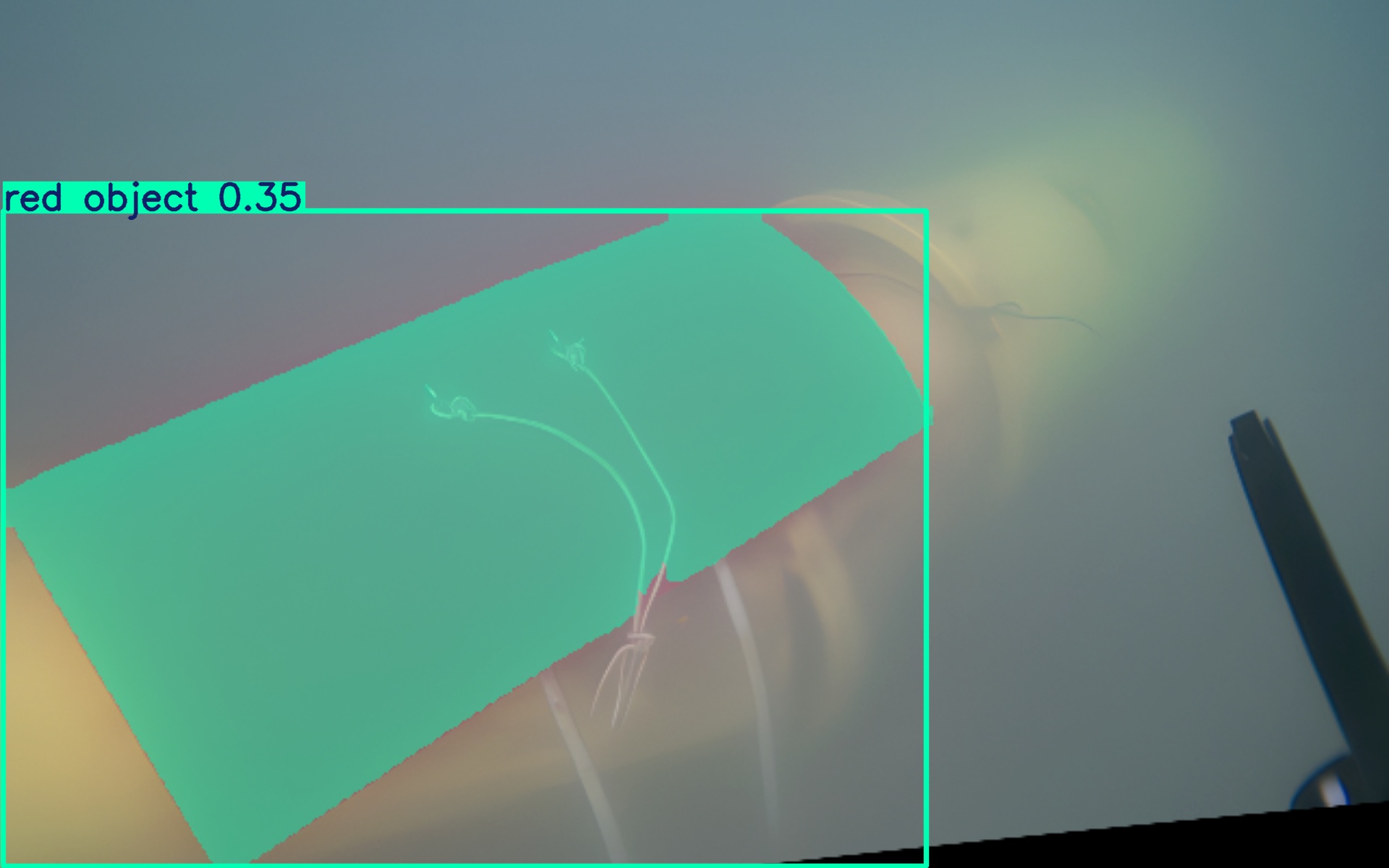} \\
    \end{tabular}
    \captionof{figure}{Visual results obtained in the left camera frame. The first column presents the rectified images, the second shows the disparity maps, and the third displays the segmented output.}
    \label{fig:stereo_results}
\end{table}

\subsection{Challenges Encountered}

On the first day of the competition, high ambient temperatures caused the Mini-Girona's internal electronics to overheat, disrupting communication with the STM board and sonar system, which impaired autonomous navigation and data acquisition for the initial two days. Additionally, NATO access restrictions at the venue limited entry only to three of the eight team members. This severe limitation imposed considerable challenges on operating the robot, managing the complex behavior trees for autonomous navigation, and executing tasks efficiently. The reduced personnel increased workload for the present team members, requiring them to cover multiple roles simultaneously and make critical decisions under immense pressure.

Despite these substantial obstacles, the team was able to secure a high overall ranking and multiple awards as a result of the platform's performance. Future iterations and deployments of the Mini-Girona will incorporate enhanced thermal management solutions and more robust contingency planning for personnel limitations to ensure more consistent autonomous performance and even greater success in future challenges.

\section{Conclusion and Future Work}

The Mini-Girona's participation in the RAMI 2025 competition, despite its recent operational status and significant challenges, demonstrated its robust design and the team's adaptability, securing second place overall and several prestigious awards. Future work will focus on enhancing the Mini-Girona's capabilities through improved perception for more reliable 3D reconstruction using deep learning, aiming for full interventional autonomy, and integrating vision into localization for more precise navigation. Additionally, the team plans to incorporate and test other sensor combinations, such as a monocular camera setup with front-facing sonar, specifically for developing and deploying advanced sensor fusion algorithms.

\section*{Acknowledgment}

The authors extend their sincere gratitude to Llu\'{i}s Mag\'{i} and Roger Pi for their extensive assistance and invaluable support throughout this project. We also thank Gabriele Ferri and the RAMI team for successfully organizing and hosting the competition.

\printbibliography

\end{document}